\RequirePackage{fix-cm}
\documentclass[a4paper,11pt]{article}

\usepackage{authblk}
\usepackage{url}
\usepackage{microtype}
\usepackage{subfigure}
\usepackage{tikz}
\usepackage{microtype}
\usepackage{xcolor}
\usepackage{multirow}
\usepackage{graphicx}
\usepackage{algorithm}
\usepackage{algpseudocode}
\usepackage{amsmath,amsfonts,amssymb}

\DeclareMathOperator*{\argmax}{arg\,max}

\newcommand{\R}{\mathbb{R}}

\DeclareMathOperator{\Cross}{Cross}
\DeclareMathOperator{\Mut}{Mut}

\providecommand{\keywords}[1]{\textbf{\textit{Keywords }} #1}

\begin{document}

\title{The Effect of Multi-Generational Selection in Geometric Semantic Genetic Programming}

\author[1]{Mauro Castelli}
\author[2]{Luca Manzoni}
\author[3]{Luca Mariot}
\author[2]{Giuliamaria Menara}
\author[2]{Gloria Pietropolli}
	
 \affil[1]{{\small NOVA IMS, Universidade NOVA de Lisboa, Campus de Campolide, 1070-312, Lisboa, Portugal}

     {\small \texttt{mcastelli@novaims.unl.pt}}}

 \affil[2]{{\small Dipartimento di Matematica e Geoscienze, Università degli Studi di Trieste, H2bis Building, Via Alfonso Valerio 12/1, 34127 Trieste, Italy}
	
 	{\small \texttt{lmanzoni@units.it}, \texttt{\{giuliamaria.menara, gloria.pietropolli\}@phd.units.it}}}

 \affil[3]{{\small Digital Security Group, Radboud University, PO Box 9010, 6500 GL Nijmegen, The Netherlands} 
	
 	{\small \texttt{luca.mariot@ru.nl}}}

\maketitle

\begin{abstract}
Among the evolutionary methods, one that is quite prominent is Genetic Programming, and, in recent years, a variant called Geometric Semantic Genetic Programming (GSGP) has shown to be successfully applicable to many real-world problems. Due to a peculiarity in its implementation, GSGP needs to store all the evolutionary history, i.e., all populations from the first one. We exploit this stored information to define a multi-generational selection scheme that is able to use individuals from older populations. We show that a limited ability to use ``old'' generations is actually useful for the search process, thus showing a zero-cost way of improving the performances of GSGP.
\end{abstract}


\section{Introduction}
\label{sec:intro}

Genetic Programming (GP)~\cite{koza1994genetic} is a method for evolving programs, usually represented as trees, through operations that mimic the Darwinian process of natural selection. Among the most successful methods of GP there is the \emph{geometric semantic GP} (GSGP)~\cite{moraglio2012geometric}. Initially, it was only an object of theoretical interest, with many interesting geometric properties on the fitness landscape that it generates. Successively, a fast way of implementing it~\cite{vanneschi2013new,castelli2019gsgp} allowed its application to many different fields (e.g.,~\cite{vanneschi2014geometric}).

To implement and execute GSGP in a fast way, it is necessary to store \emph{all} information regarding the entire evolutionary process. In particular, all the populations from the first one need to be stored efficiently and compactly. Therefore, it is only natural to ask \emph{how} this information can be used to improve the search process.

In this paper, we propose a general way of using this additional ``free'' information by allowing the selection process to select individuals from ``old'' populations. In particular, we present two ways of performing this multi-generational selection:
\begin{itemize}
    \item By selecting uniformly among the last $k$ generations;
    \item By selecting among all generations with a decreasing probability (i.e., with a geometric distribution).
\end{itemize}
We compare the two methods with multiple parameters on six different real-world datasets, showing that a limited ability to look back on the evolutionary history of GSGP is actually beneficial in the evolutionary process.

This paper is structured as follows: Section~\ref{sec:related} explores the existing works on using old generations (or \emph{memory}) to improve evolutionary computation algorithms. Section~\ref{sec:multi-generational-selection} recalls the basic notions of GSGP and introduces multi-generational selection. In Section~\ref{sec:experimental-settings} the experimental settings are presented, while in Section~\ref{sec:results-and-discussion} the results are presented and discussed. Section~\ref{sec:conclusions} provides a summary of this work and directions for future research.

\section{Related Works}
\label{sec:related}
When considering an evolutionary algorithm (EA) as a dynamical system, the idea of selecting individuals from older populations may be regarded as the ability of the system to directly exploit its \emph{memory}, or equivalently the sequence of its past states. Under this perspective, approaches that enhanced Genetic Algorithms (GA) with memory started to appear in the literature already from the mid 90s.

As far as we know, Louis and Li~\cite{louis97} were the first to propose the use of a \emph{long-term memory} to store the best solutions found so far by a GA, eventually reintegrating them into the population in a later stage. Their experimental investigation over the Traveling Salesman Problem (TSP) showed that GA obtained a better performance when its population was seeded with sub-optimal solutions found in previous instances, rather than initializing it at random. Wiering~\cite{wiering04} experimented with a combination of GA and local search where memory plays an analogous role in Tabu Search: if a local optimum has been found before, then it gets the lowest possible fitness to maximize the probability that the corresponding individual is replaced in the next generation. Later, Yang~\cite{yang08} compared two variants of the random immigrants scheme respectively based on memory and elitism, with the goal of enhancing the performance of GA over dynamic fitness landscapes. In these hybrid schemes, the best individual retained either by memory or elitism from old populations was used as a basis to evolve new immigrants through mutation, therefore increasing the diversity of the population and its adaptability against a dynamic environment. Still with regard to dynamic optimization problems, Cao and Luo~\cite{cao09} considered two retrieving strategies which selected the two best individuals from the associative memory of a GA. In particular, the environmental information associated with these two individual was evaluated by either a survivability or diversity criterion. Similar to the methods proposed in this paper are the previous works in reinsertion of old genetic material in GA by Castelli et al.~\cite{castelli2011effect,castelli2011method}. There, the authors proposed a method to boost the GA optimization ability by replacing a fraction of the worst individuals with the best ones from an older population.

From the point of view of GP, the reinsertion of genetic material from old populations usually occurs in the related literature under the name of \emph{concept and knowledge reuse}. This is indeed a proper term, since GP evolves programs that can be used in turn for learning concepts and functions, as e.g. in symbolic regression~\cite{augusto00}. Séront~\cite{seront95} set forth a method to retrieve concepts evolved by GP based on a library that saved the trees of the best individuals. This method stood on the reasonable assumption that highly fit individuals embed useful concepts in their syntactic trees for solving a particular optimization problem. The results showed that the use of a concept library to create the initial population is beneficial for GP, as compared to random initialization. Jaskowski et al.~\cite{jaskowski07} explored a different direction where a method for reusing knowledge embedded by GP was used among a group of learners that worked in parallel on a visual learning in task. Therefore, in this case the reuse of GP subprograms does not come from old populations, but is rather shared among different current populations at the same time. Pei et al.~\cite{pei19} investigated the issue of class imbalance in GP-based classifiers, and proposed a method to mitigate it by using previously evolved GP trees to initialize the population in later runs. The experimental results indicated that such mechanism allows to reduce the training time and increase the accuracy of multi-classifier systems based on GP. More recently, Bi et al.~\cite{bi21} proposed a new method to improve GP learning performance over image classification problems Such a method is based on knowledge transfer among multiple populations, similarly to the aforementioned approach of~\cite{jaskowski07}.

One of the main issues of the methods proposed in the above papers is that (a part of) the evolutionary history is needed to properly exploit older populations, thus increasing the space necessary for those methods to work. It is also interesting to notice that an increase in the populations' size (under a certain limit) is useful also for GSGP~\cite{castelli2017influence}. Thus, as proposed in this paper, it is fundamental to explore the trade-off between population size and performance and whether this trade-off can actually be removed or mitigated by using part of the existing evolutionary history.

Remark that the principle of exploiting memory in evolutionary algorithms is also considered under a different guise in the area of machine learning. This is the case, for instance, of the \emph{conservation machine learning} approach proposed by Sipper and Moore~\cite{sipper20,sipper21}. There, the authors explore the idea of reusing ML models learned in different ways (e.g. multiple runs, ensemble methods, etc.) and apply it to the case of random forests. The results showed that their method improves the performance over certain classification tasks through ensemble cultivation.

Finally, notice that in the specific area of GSGP there is no work addressing directly the reuse of old genetic material or individuals from past populations, to the best of our knowledge. Nevertheless, there have been several attempts aimed at improving the performance of the basic GSGP algorithm over regression problems. One of the most developed research lines in this respect considers for instance \emph{local search}, which is used as an additional optimization step combined with the evolutionary process of GSGP~\cite{castelli15,castelli19}. 

\section{Geometric Semantic GP with Multi-Generational Selection}
\label{sec:GSGP-with-MGS}

In this section, we recall the construction of GSGP with the semantic operators and the current efficient implementation through dynamic programming. We then introduce a procedure to use this implementation to obtain a zero-cost way of sampling, during the selection process, not only from the current population but also for \emph{any} previous population obtained during the evolution process.

\subsection{Geometric Semantic Genetic Programming}
\label{sec:GSGP}

In classical GP, the crossover (recombination) and mutation operators act on the genotype of the individual involved, usually modifying or changing parts of the subtrees in the case of a tree-based representation. However, the effects on the phenotype of the individuals are difficult to formalize, and even small modifications to the genotype can create a significant change in the phenotype. Therefore, in addition to the standard ``syntactic'' operators of GP, there have been multiple studies on \emph{semantic} operators~\cite{vanneschi2014survey}. By relying on these operators, it is possible to predict the effect of crossover and mutation on the phenotype of the solutions.

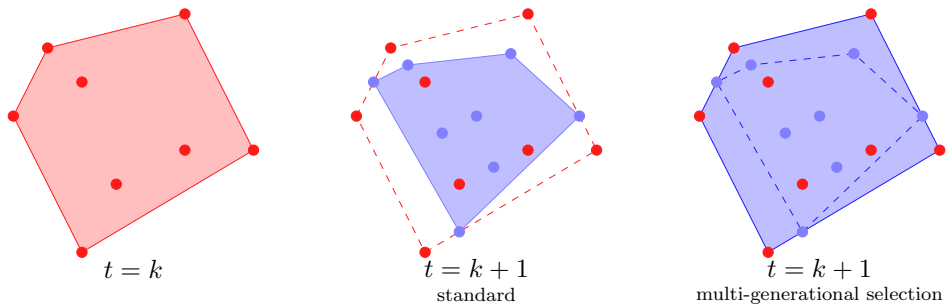
\begin{figure}
    \centering
    \resizebox{\textwidth}{!}{
    \begin{tikzpicture}[scale=0.5]
        \filldraw[color=red!90,fill=red!25] (0,4) -- (1,6) -- (5,7) -- (7,3) -- (2,0) -- (0,4);

        \filldraw[color=red!90] (0,4) circle (0.15cm);
        \filldraw[color=red!90] (2,0) circle (0.15cm);
        \filldraw[color=red!90] (1,6) circle (0.15cm);
        \filldraw[color=red!90] (5,7) circle (0.15cm);
        \filldraw[color=red!90] (7,3) circle (0.15cm);
    
        \filldraw[color=red!90] (3,2) circle (0.15cm);
        \filldraw[color=red!90] (2,5) circle (0.15cm);
        \filldraw[color=red!90] (5,3) circle (0.15cm);
        
        \draw (3.5, -0.5) node {$t=k$};
    
        \draw[color=red!90,dashed] (10,4) -- (11,6) -- (15,7) -- (17,3) -- (12,0) -- (10,4);
        \filldraw[color=blue!50,fill=blue!25] (13,3/5) -- (16.5,4) -- (14.5,35/6) -- (11.5,5.5) -- (10.5,5) -- (13,3/5);

        \filldraw[color=red!90] (10,4) circle (0.15cm);
        \filldraw[color=red!90] (12,0) circle (0.15cm);
        \filldraw[color=red!90] (11,6) circle (0.15cm);
        \filldraw[color=red!90] (15,7) circle (0.15cm);
        \filldraw[color=red!90] (17,3) circle (0.15cm);
    
        \filldraw[color=red!90] (13,2) circle (0.15cm);
        \filldraw[color=red!90] (12,5) circle (0.15cm);
        \filldraw[color=red!90] (15,3) circle (0.15cm);
    
        \filldraw[color=blue!50] (10.5,5) circle (0.15cm);
        \filldraw[color=blue!50] (11.5,5.5) circle (0.15cm);
        \filldraw[color=blue!50] (16.5,4) circle (0.15cm);
        \filldraw[color=blue!50] (13,3/5) circle (0.15cm);
        \filldraw[color=blue!50] (14.5,35/6) circle (0.15cm);

        \filldraw[color=blue!50] (12.5,3.5) circle (0.15cm);
        \filldraw[color=blue!50] (14,2.5) circle (0.15cm);
        \filldraw[color=blue!50] (13.5,4) circle (0.15cm);
        
        \draw (13.5, -0.5) node {$t=k+1$};
        \draw (13.5, -1.25) node {\scriptsize standard};
        
        \filldraw[color=blue!90,fill=blue!25] (20,4) -- (21,6) -- (25,7) -- (27,3) -- (22,0) -- (20,4);
        \draw[color=blue!90,dashed] (23,3/5) -- (26.5,4) -- (24.5,35/6) -- (21.5,5.5) -- (20.5,5) -- (23,3/5);

        \filldraw[color=red!90] (20,4) circle (0.15cm);
        \filldraw[color=red!90] (22,0) circle (0.15cm);
        \filldraw[color=red!90] (21,6) circle (0.15cm);
        \filldraw[color=red!90] (25,7) circle (0.15cm);
        \filldraw[color=red!90] (27,3) circle (0.15cm);
    
        \filldraw[color=red!90] (23,2) circle (0.15cm);
        \filldraw[color=red!90] (22,5) circle (0.15cm);
        \filldraw[color=red!90] (25,3) circle (0.15cm);
    
        \filldraw[color=blue!50] (20.5,5) circle (0.15cm);
        \filldraw[color=blue!50] (21.5,5.5) circle (0.15cm);
        \filldraw[color=blue!50] (26.5,4) circle (0.15cm);
        \filldraw[color=blue!50] (23,3/5) circle (0.15cm);
        \filldraw[color=blue!50] (24.5,35/6) circle (0.15cm);

        \filldraw[color=blue!50] (22.5,3.5) circle (0.15cm);
        \filldraw[color=blue!50] (24,2.5) circle (0.15cm);
        \filldraw[color=blue!50] (23.5,4) circle (0.15cm);
        
        \draw (23.5, -0.5) node {$t=k+1$};
        \draw (23.5, -1.25) node {\scriptsize multi-generational selection};

    \end{tikzpicture}
    }
    \caption{A representation of the effect of multi-generational selection on the convex hull where geometric semantic crossover can generate new individuals.}
    \label{fig:convex-hull-population}
\end{figure}

In 2012 Moraglio and coauthors~\cite{moraglio2012geometric} defined a way of performing semantic crossover and mutation where the effect was obtaining a \emph{geometric} crossover and \emph{geometric} mutation in the semantic space, two concepts defined and studied in~\cite{moraglio2004topological}. Given a metric space $(X, d)$ where $X$ is a set and $d \colon X \times X \to \R_+$ is a distance, a crossover is said to be geometric if, for each $x,y \in X$, each element resulting from the crossover of $x$ and $y$ belongs to the set $\{z \in A \;|\; d(x,z) + d(z,y) = d(x,y)\}$, i.e., the result of the crossover is inside the segment connecting $x$ and $y$. A mutation operator is said to be geometrical and, in particular, a geometric $\varepsilon$-mutation, if, for each $x \in X$, the individuals resulting from the mutation are all inside the ball $B_\varepsilon(x)$, i.e., the ball of radius $\varepsilon$ centered in $x$. One of the advantages of working with geometric operators is that it is possible to reason on the geometrical effect of the different operators; e.g., a geometric crossover always generates offspring in the convex hull given by the current population (see Figure~\ref{fig:convex-hull-population}).

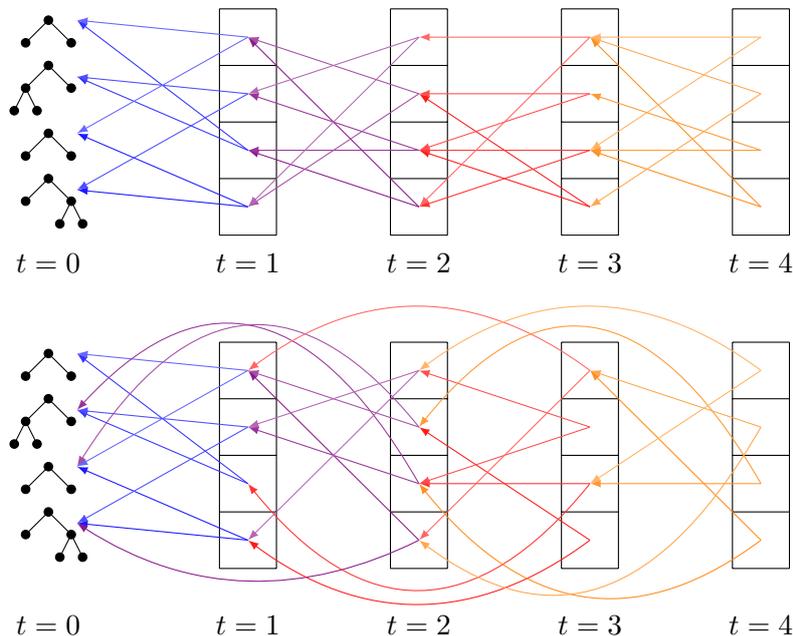
\begin{figure}
    \centering
    \begin{tikzpicture}[scale=0.75]

        \draw (-3,1) -- (-3.4,0.6);
        \draw (-3,1) -- (-2.6,0.6);
        \draw (-2.6,0.6) -- (-2.8,0.2);
        \draw (-2.6,0.6) -- (-2.4,0.2);
        \filldraw (-3,1) circle (0.075cm);
        \filldraw (-3.4,0.6) circle (0.075cm);
        \filldraw (-2.6,0.6) circle (0.075cm);
        \filldraw (-2.8,0.2) circle (0.075cm);
        \filldraw (-2.4,0.2) circle (0.075cm);

        \draw (-3,1.8) -- (-3.4,1.4);
        \draw (-3,1.8) -- (-2.6,1.4);
        \filldraw (-3,1.8) circle (0.075cm);
        \filldraw (-3.4,1.4) circle (0.075cm);
        \filldraw (-2.6,1.4) circle (0.075cm);

        \draw (-3,3) -- (-3.4,2.6);
        \draw (-3,3) -- (-2.6,2.6);
        \draw (-3.4,2.6) -- (-3.2,2.2);
        \draw (-3.4,2.6) -- (-3.6,2.2);
        \filldraw (-3,3) circle (0.075cm);
        \filldraw (-3.4,2.6) circle (0.075cm);
        \filldraw (-2.6,2.6) circle (0.075cm);
        \filldraw (-3.2,2.2) circle (0.075cm);
        \filldraw (-3.6,2.2) circle (0.075cm);
        
        \draw (-3,3.8) -- (-3.4,3.4);
        \draw (-3,3.8) -- (-2.6,3.4);
        \filldraw (-3,3.8) circle (0.075cm);
        \filldraw (-3.4,3.4) circle (0.075cm);
        \filldraw (-2.6,3.4) circle (0.075cm);
    
        \draw[step=1cm] (0,0) grid (1,4); 
        \draw[step=1cm] (3,0) grid (4,4); 
        \draw[step=1cm] (6,0) grid (7,4); 
        \draw[step=1cm] (9,0) grid (10,4); 
        
        \draw (-3,-0.5) node  {$t=0$};
        \draw (0.5,-0.5) node  {$t=1$};
        \draw (3.5,-0.5) node  {$t=2$};
        \draw (6.5,-0.5) node  {$t=3$};
        \draw (9.5,-0.5) node  {$t=4$};
    
        \draw[-latex,color=blue!90] (0.5,0.5) -- (-2.5,1.8);
        \draw[-latex,color=blue!90] (0.5,0.5) -- (-2.5,0.8);
        \draw[-latex,color=blue!80] (0.5,1.5) -- (-2.5,2.8);
        \draw[-latex,color=blue!80] (0.5,1.5) -- (-2.5,3.8);
        \draw[-latex,color=blue!70] (0.5,2.5) -- (-2.5,0.8);
        \draw[-latex,color=blue!70] (0.5,2.5) -- (-2.5,2.8);
        \draw[-latex,color=blue!60] (0.5,3.5) -- (-2.5,1.8);
        \draw[-latex,color=blue!60] (0.5,3.5) -- (-2.5,3.8);
    
        \draw[-latex,color=violet!90] (3.5,0.5) -- (0.5,1.5);
        \draw[-latex,color=violet!90] (3.5,0.5) -- (0.5,3.5);
        \draw[-latex,color=violet!80] (3.5,1.5) -- (0.5,1.5);
        \draw[-latex,color=violet!80] (3.5,1.5) -- (0.5,2.5);
        \draw[-latex,color=violet!70] (3.5,2.5) -- (0.5,0.5);
        \draw[-latex,color=violet!70] (3.5,2.5) -- (0.5,3.5);
        \draw[-latex,color=violet!60] (3.5,3.5) -- (0.5,0.5);
        \draw[-latex,color=violet!60] (3.5,3.5) -- (0.5,2.5);
        
        \draw[-latex,color=red!90] (6.5,0.5) -- (3.5,2.5);
        \draw[-latex,color=red!90] (6.5,0.5) -- (3.5,1.5);
        \draw[-latex,color=red!80] (6.5,1.5) -- (3.5,0.5);
        \draw[-latex,color=red!80] (6.5,1.5) -- (3.5,1.5);
        \draw[-latex,color=red!70] (6.5,2.5) -- (3.5,1.5);
        \draw[-latex,color=red!70] (6.5,2.5) -- (3.5,2.5);
        \draw[-latex,color=red!60] (6.5,3.5) -- (3.5,3.5);
        \draw[-latex,color=red!60] (6.5,3.5) -- (3.5,0.5);
        
        \draw[-latex,color=orange!90] (9.5,0.5) -- (6.5,1.5);
        \draw[-latex,color=orange!90] (9.5,0.5) -- (6.5,3.5);
        \draw[-latex,color=orange!80] (9.5,1.5) -- (6.5,1.5);
        \draw[-latex,color=orange!80] (9.5,1.5) -- (6.5,2.5);
        \draw[-latex,color=orange!70] (9.5,2.5) -- (6.5,3.5);
        \draw[-latex,color=orange!70] (9.5,2.5) -- (6.5,0.5);
        \draw[-latex,color=orange!60] (9.5,3.5) -- (6.5,3.5);
        \draw[-latex,color=orange!60] (9.5,3.5) -- (6.5,1.5);
    \end{tikzpicture}
    
    \begin{tikzpicture}[scale=0.75]

        \draw (-3,1) -- (-3.4,0.6);
        \draw (-3,1) -- (-2.6,0.6);
        \draw (-2.6,0.6) -- (-2.8,0.2);
        \draw (-2.6,0.6) -- (-2.4,0.2);
        \filldraw (-3,1) circle (0.075cm);
        \filldraw (-3.4,0.6) circle (0.075cm);
        \filldraw (-2.6,0.6) circle (0.075cm);
        \filldraw (-2.8,0.2) circle (0.075cm);
        \filldraw (-2.4,0.2) circle (0.075cm);

        \draw (-3,1.8) -- (-3.4,1.4);
        \draw (-3,1.8) -- (-2.6,1.4);
        \filldraw (-3,1.8) circle (0.075cm);
        \filldraw (-3.4,1.4) circle (0.075cm);
        \filldraw (-2.6,1.4) circle (0.075cm);

        \draw (-3,3) -- (-3.4,2.6);
        \draw (-3,3) -- (-2.6,2.6);
        \draw (-3.4,2.6) -- (-3.2,2.2);
        \draw (-3.4,2.6) -- (-3.6,2.2);
        \filldraw (-3,3) circle (0.075cm);
        \filldraw (-3.4,2.6) circle (0.075cm);
        \filldraw (-2.6,2.6) circle (0.075cm);
        \filldraw (-3.2,2.2) circle (0.075cm);
        \filldraw (-3.6,2.2) circle (0.075cm);
        
        \draw (-3,3.8) -- (-3.4,3.4);
        \draw (-3,3.8) -- (-2.6,3.4);
        \filldraw (-3,3.8) circle (0.075cm);
        \filldraw (-3.4,3.4) circle (0.075cm);
        \filldraw (-2.6,3.4) circle (0.075cm);
    
        \draw[step=1cm] (0,0) grid (1,4); 
        \draw[step=1cm] (3,0) grid (4,4); 
        \draw[step=1cm] (6,0) grid (7,4); 
        \draw[step=1cm] (9,0) grid (10,4); 
        
        \draw (-3,-1) node  {$t=0$};
        \draw (0.5,-1) node  {$t=1$};
        \draw (3.5,-1) node  {$t=2$};
        \draw (6.5,-1) node  {$t=3$};
        \draw (9.5,-1) node  {$t=4$};
    
        \draw[-latex,color=blue!90] (0.5,0.5) -- (-2.5,1.8);
        \draw[-latex,color=blue!90] (0.5,0.5) -- (-2.5,0.8);
        \draw[-latex,color=blue!80] (0.5,1.5) -- (-2.5,2.8);
        \draw[-latex,color=blue!80] (0.5,1.5) -- (-2.5,3.8);
        \draw[-latex,color=blue!70] (0.5,2.5) -- (-2.5,0.8);
        \draw[-latex,color=blue!70] (0.5,2.5) -- (-2.5,2.8);
        \draw[-latex,color=blue!60] (0.5,3.5) -- (-2.5,1.8);
        \draw[-latex,color=blue!60] (0.5,3.5) -- (-2.5,3.8);
    
        \draw[-latex,color=violet!90] (3.5,0.5) .. controls (1.5,-0.5) and (-0.5,-0.5) .. (-2.5,0.8);
        \draw[-latex,color=violet!90] (3.5,0.5) -- (0.5,3.5);
        \draw[-latex,color=violet!80] (3.5,1.5) .. controls (1.5,5) and (-0.5,5) ..  (-2.5,2.8);
        \draw[-latex,color=violet!80] (3.5,1.5) -- (0.5,2.5);
        \draw[-latex,color=violet!70] (3.5,2.5) .. controls (1.5,5) and (-0.5,5) .. (-2.5,1.8);
        \draw[-latex,color=violet!70] (3.5,2.5) -- (0.5,3.5);
        \draw[-latex,color=violet!60] (3.5,3.5) -- (0.5,0.5);
        \draw[-latex,color=violet!60] (3.5,3.5) -- (0.5,2.5);
        
        \draw[-latex,color=red!90] (6.5,0.5) -- (3.5,2.5);
        \draw[-latex,color=red!90] (6.5,0.5) .. controls (4.5,-1) and (2.5,-1) .. (0.5,0.5);
        \draw[-latex,color=red!80] (6.5,1.5) .. controls (4.5,-1) and (2.5,-1) .. (0.5,1.5);
        \draw[-latex,color=red!80] (6.5,1.5) -- (3.5,1.5);
        \draw[-latex,color=red!70] (6.5,2.5) -- (3.5,3.5);
        \draw[-latex,color=red!70] (6.5,2.5) -- (3.5,1.5);
        \draw[-latex,color=red!60] (6.5,3.5) -- (3.5,0.5);
        \draw[-latex,color=red!60] (6.5,3.5) .. controls (4.5,5) and (2.5,5) .. (0.5,3.5);
        
        \draw[-latex,color=orange!90] (9.5,0.5) .. controls (7.5,-1) and (5.5,-1) .. (3.5,1.5);
        \draw[-latex,color=orange!90] (9.5,0.5) -- (6.5,3.5);
        \draw[-latex,color=orange!80] (9.5,1.5) -- (6.5,1.5);
        \draw[-latex,color=orange!80] (9.5,1.5) .. controls (7.5,5) and (5.5,5) .. (3.5,2.5);
        \draw[-latex,color=orange!70] (9.5,2.5) -- (6.5,3.5);
        \draw[-latex,color=orange!70] (9.5,2.5) .. controls (7.5,-1) and (5.5,-1) .. (3.5,0.5);
        \draw[-latex,color=orange!60] (9.5,3.5) .. controls (7.5,5) and (5.5,5) ..  (3.5,3.5);
        \draw[-latex,color=orange!60] (9.5,3.5) -- (6.5,1.5);
    \end{tikzpicture}
    \caption{A visual representation of how GSGP can be implemented in an efficient way, sharing subtrees between individuals. At the top, the standard implementation where the parents can be selected only from the previous generation. At the bottom, parents can be selected uniformly at random from the previous two populations. Notice how no additional storage is required.}
    \label{fig:GSGP-pointers}
\end{figure}

In particular, one interesting space where it is possible to define genetic operators is the \emph{semantic space}. That is, given a set of samples $X = \{\vec{x}^1, \ldots, \vec{x}^m\} \subset \R^n$ and a function $T : \R^n \to \R$, representing a GP individual, we can define the \emph{semantics of $T$}, denoted by $s(T)$ as the vector $(T(\vec{x}^1), T(\vec{x}^2), \ldots, T(\vec{x}^m)) \in \R^m$. If we perform a \emph{geometric} crossover between the semantics of two trees $T_1$ and $T_2$, the semantics of \emph{any} individual resulting from the crossover will have outputs (on the elements of the set $X$) intermediate between the ones of $T_1$ and $T_2$. Similarly, a \emph{geometric} mutation on the semantic space will generate a perturbation of the outputs of the parent individual. There is, however, the question of how to define operators working on the semantic space by manipulating the syntactic space. The conundrum was solved by Moraglio et Al.~\cite{moraglio2012geometric} in 2012 with the definition of semantic mutation and crossover, as detailed below. 

\textbf{Semantic Crossover}. Let $T_1$ and $T_2$ be two functions from $\R^n$ to $\R$ representing two GP trees and let $R \colon \R^n \to [0,1]$ be a randomly generated tree. Then the \emph{semantic crossover} between $T_1$ and $T_2$ using the random tree $R$ is defined as:
\begin{align*}
    & \Cross(T_1, T_2, R)(\vec{x}) = R(\vec{x}) T_1(\vec{x}) + (1-R(\vec{x}))T_2(\vec{x}) & \text{for all $\vec{x} \in \R^n$.}
\end{align*}
Notice that the outputs of $\Cross(T_1, T_2, R)(\vec{x})$ will be intermediate with respect to the outputs of $T_1$ and $T_2$. Thus, the crossover is geometric in the semantic space. To constrain the output of $R$ inside the interval $[0,1]$, a simple technique is to generate a random tree and pass its output to a sigmoid function.

\textbf{Semantic Mutation}. Let $T \colon \R^n \to \R$ be the function defined by a GP tree, $R \colon \R^n \to \R$ be a randomly generated tree, and $m \in \R_+$ be a positive real number, called the \emph{mutation step}. Then, the \emph{semantic mutation} of $T$ using the random tree $R$ is defined as:
\begin{align*}
    & \Mut(T,R)(\vec{x}) = T(\vec{x}) + m R(\vec{x}) & \text{for all $\vec{x} \in \R^n$.}
\end{align*}
Notice that there is an additional parameter, the mutation step $m$, that allows tuning how ``big'' are the jumps/perturbations produced by mutation.

One disadvantage of geometric operators and crossover in particular -- at least in their na\"ive implementation -- is that they produce trees that are exponential with respect to the number of generations: each crossover can be more than twice the size of the trees, thus generating an exponential blowup of the trees. This problem can be solved by using a dynamic programming approach~\cite{vanneschi2013new}:
\begin{itemize}
    \item The initial population is composed of standard GP trees;
    \item Each successive generation is not composed of trees; rather, each individual is a structure containing the random trees used in crossover and mutations and \emph{pointers} or \emph{references} to the individuals in the previous populations. This solves the problem of an exponential space blowup;
    \item Evaluation can be performed bottom-up, saving the intermediate results from the initial population and combining them following the application of crossover and mutations.
\end{itemize}
The resulting structure is represented schematically in Figure~\ref{fig:GSGP-pointers}. The resulting complexity for computing the output of an individual is $O(g p)$ where $g$ is the number of generations, and $p$ is the population size. Thus, the resulting complexity is polynomial rather than exponential.

\subsection{Multi-Generational Selection for GSGP}
\label{sec:multi-generational-selection}

One interesting aspect to notice in the ``fast'' GSGP implementation is that we are actually able to access all the intermediate populations at zero cost: they are already stored in the data structures that we use, and their output is already computed to obtain the outputs of the current population. Hence, it is natural to ask if we can use this additional ``free'' information to improve the performances in GSGP. In particular, we propose to allow ``older'' individuals to be selected in the tournament selection. 

Concerning the implementation details, as one can observe in Figure~\ref{fig:GSGP-pointers} at the bottom, the pointers can go back for multiple generations (two in the example figure) instead of a single one, with no additional overhead. The pseudocode for the multi-generational selection is presented in Algorithm~\ref{alg:pseudocode}. As it is possible to notice, since all the information is already available, there is no significant additional computational cost compared to standard GSGP.

\begin{algorithm}
    \begin{algorithmic}
        \Function{Multi-generational-selection}{$P$, $n$, $f$, $t$, $D$}
                        \State Tournament $\gets \varnothing$\; \Comment{Individuals selected for the tournament}
            \For{$1 \le i \le t$} \Comment{Repeat for the tournament size $t$}
                \State $j \gets n - \text{ extract from }D$\; \Comment{Select the generation}
                \State $k \gets \text{ uniform random integer between $1$ and $|P[j]|$}$ \; \Comment{Select the individual}
                \State Tournament $\gets$ Tournament $\cup \{P[j][k]\}$\; \Comment{Add the individual to the tournament}
            \EndFor
            \State best $\gets \argmax_{x \in \text{Tournament}} f(x)$\; \Comment{Find the best individual in the tournament}
            \State \Return best\;
        \EndFunction
    \end{algorithmic}
    \caption{\label{alg:pseudocode}The pseudocode of the multi-generational (tournament) selection algorithm, where $P$ is a two-dimensional array of individuals of $n$ rows (generations) where $P[i][j]$ is the $j$-th individual in the $i$-th generation, $f$ is the fitness function, $t \in \mathbb{N}$ is the tournament size, and $D$ is a distribution.}
\end{algorithm}

In what follows, we describe how this selection can be performed since multiple possibilities are available. In particular, we propose a uniform and a geometric selection strategy.

\subsubsection{Uniform Multi-Generational Selection}

The simplest case is when individuals can be selected \emph{uniformly at random} from the populations of the last $k \in \mathbb{N}$ generations (or all of them if less than $k$ generations have been performed). If $k=1$, then we obtain standard GSGP. However, as shown in Figure~\ref{fig:convex-hull-population}, one of the effects of higher values of $k$ is to ``delay'' the shrinking of the convex hull given by crossover. In this way, we expect to equip GSGP with a better exploration ability.

\subsubsection{Geometric Multi-Generational Selection}

In the uniform selection method, there is a hard cut-off for participating in the selection process. Another idea is to gradually decrease the probability of an old population contributing to the selection process. In particular, it is possible to employ a \emph{geometric distribution}, so that the probability of selecting the $k$-th generation before the current one is $p(1-p)^k$ with $p \in (0,1)$ a parameter. Thus, for example, if $k = 0.5$, the probability of selecting from the previous population is $0.5$, two generations behind $(0.5)^2 = 0.25$, and so on (if we go back more than the number of existing generations we select from the initial population). Thus, we can regard the geometric method as a way of ``fading out'' the convex hull given by the previous populations.

Here we have defined two different multi-generations selection methods: \emph{uniform} with a parameter $k \in \mathbb{N}$, which we will denote by U$k$ (e.g., U$1$, U$2$, etc.), and \emph{geometric} with a parameter $p \in (0,1)$, which we will denote by G$p$ (e.g., G$0.5$, G$0.75$, etc.).

\section{Experimental Setting}
\label{sec:experimental-settings}
The following section introduces the experimental environment adopted to test the validity of the proposed methods: Section~\ref{subsec:dataset} describes the benchmark dataset employed, while Section~\ref{subsec:experimental-study} provides the experimental settings needed to render experiments fully reproducible.

\subsection{Dataset}
\label{subsec:dataset}
The datasets exploited in this paper have been purposely chosen as they consist of real-world, complex datasets ranging from different areas that have been extensively leveraged as benchmarks for GP. The reader can find \cite{mcdermott2012genetic} a comprehensive description of the reasons why these datasets represent suitable reference points to assert the validity of a genetic programming method.
Table~\ref{tab:dataset} outlines the key features of the considered problems, such as the number of instances and variables, the area to which they belong, and, finally, the kind of task required. It is worth pointing out that these datasets are significantly different from each other (considering, for example, the number of instances and variables). Thus, they represent an optimal choice to test the validity of our methods when applied to problems that have rather diverse characteristics. 

The first group of datasets deals with predicting pharmacokinetic parameters of potential new drugs.

\textit{Human oral bioavailability} (\textbf{\%F}) measures the percentage of initial drug dose that effectively reaches the systemic blood circulation: this problem constitutes an essential pharmacokinetic task as the oral assumption is usually the preferred way of supplying drugs to patients, and also because it is a representative measure of the quantity of active principle that can effectively actuate its biological effect \cite{archetti2007genetic}. 

\textit{Protein-plasma binding level} (\textbf{\%PPB}) characterizes the distribution into the human body of a drug. Specifically, it corresponds to the percentage of the initial drug dose which binds plasma proteins: this measure is fundamental, as blood circulation is the major vehicle of drug distribution into the human body \cite{archetti2006genetic}.

\textit{Median Oral Lethal Dose} (\textbf{LD50}) concerns the harmful effect produced by the distribution of a drug into the human body, as it measures the lethal dose required to kill half the members of a tested population after a specified time. It is expressed as the number of milligrams of drug-related to one kilogram of cavies mass \cite{archetti2006genetic}.

The second group of datasets originates from physical problems. 

\textit{Airfoil self-noise} (\textbf{air}) measures the hydrodynamic performance of sailing yachts taking into account their dimension and velocity \cite{brooks1989airfoil}.

\textit{Concrete compressive strength} (\textbf{conc})  \cite{castelli2013prediction} characterizes the value of the slump flow of the concrete when given as inputs concrete components such as cement, fly ash, slag, water, coarse aggregate and fine aggregate.

Finally, \textit{Yacht hydrodynamics} (\textbf{yac}) measures the hydrodynamic performance of sailing yachts starting from their dimension and velocity.

\begin{table}[h]
    \centering
    \scriptsize
    \begin{tabular}{ l | c | c | c | c }
         \textbf{Dataset} & \textbf{Variables} & \textbf{Instances}
         & \textbf{Area} \\
         \hline
         airfoil & 6 & 1503 & Physics \\
         bioav & 242 & 359 & Pharmacokinetic \\
         concrete & 9 & 1030 & Physics \\
         ppb & 627 & 131 & Pharmacokinetic \\
         toxicity & 627 & 234 & Pharmacokinetic \\
         yacht & 7 & 308 & Physics \\
    \end{tabular}
    \caption{Principal characteristics of the considered datasets: the number of variables, the number of instances, and the domain.}
\label{tab:dataset}
\end{table}

\subsection{Experimental Study}
\label{subsec:experimental-study}

\begin{table}[h]
    \centering 
    \scriptsize
    \begin{tabular}{ l c | c c }
         \textbf{Parameter} & $ \ $ & $ \ $ & \textbf{Value}  \\
         \hline
         Population size &  &  & $100$  \\
         Number of generations &  &  & $100$  \\
         Number of runs &  &  & $100$  \\
         Max. initial depth &  &  & $4$  \\
         Crossover rate &  & &  $0.9$  \\
         Mutation rate &  & & $0.3$ \\
         Mutation step &  & & $0.1$ \\
         Selection method &  & & Tournament of size $4$ \\
         Elitism & & & Best individuals survive \\
    \end{tabular}
    \caption{Experimental settings.}
\label{tab:parameters}
\end{table}

This section describes the experimental settings summarized in Table~\ref{tab:parameters} to make the method fully reproducible.
To get statistically valuable results, $100$ runs of the method have been performed for each benchmark considered in order to be have enough run to perform statistical tests each one consisting of $100$ generations in order to allow the algorithm to stabilize. In each run, the dataset has been randomly split into training and test sets with a percentage equal to $70\%-30\%$. All the parameters of GSGP described in Table~\ref{tab:parameters} are the standard values used in the literature (e.g.,~\cite{pietropolli2022combining}).

To assess the validity of both uniform and geometric distributions for performing the multi-generational selection, the fitness values achieved within these techniques are compared with classical GSGP. It is worth emphasizing that the results obtained by standard GP~\cite{koza1994genetic} are not listed in this paper since standard GP is consistently outperformed by GSGP. 

Regarding the uniform distribution variation of the proposed method, experiments have been performed considering the following values as the number of previous generations from where parents have been selected: $2$, $5$, $10$, $20$, and $100$. The geometric variation of the proposed technique has been tested by selecting $0.25$, $0.50$, and $0.75$ as values for the parameter $p$ that defines the geometric probability distributions. The selection of those parameters allows to cover most cases (from ``looking back'' only a few generations to the entire evolution), thus giving a general idea of the behaviour of the proposed method.

The population size for all the considered cases of study is set to $100$, which is a usual trade-off between computational costs and quality of the search process~\cite{castelli2017influence}, and the trees of the first generation are initialized with the ramped half and half technique.
The fitness function selected to measure and compare the quality of different methods proposed is the Root Mean Squared Error (RMSE).

\section{Results and Discussion}
\label{sec:results-and-discussion}

\begin{table}[]
    \scriptsize
    \centering
    \begin{tabular}{r|lllllllll|}
    & U2 & U5 & U10 & U20 & U50 & U100 & G0.25 & G0.50 & G0.75 \\
    \hline
    	airfoil & $0.158$ & $\mathbf{0.000}$ & $\mathbf{0.000}$ & $\mathbf{0.000}$ & $0.280$ & $0.000$ & $\mathbf{0.000}$ & $0.000$ & $0.000$ \\
    	bioav & $0.741$ & $\mathbf{0.000}$ & $\mathbf{0.001}$ & $\mathbf{0.000}$ & $0.000$ & $0.000$ & $\mathbf{0.007}$ & $0.000$ & $0.000$ \\
    	concrete & $0.763$ & $\mathbf{0.001}$ & $\mathbf{0.042}$ & $0.445$ & $0.001$ & $0.000$ & $0.557$ & $0.000$ & $0.000$ \\
    	ppb & $0.000$ & $\mathbf{0.000}$ & $\mathbf{0.000}$ & $\mathbf{0.000}$ & $0.000$ & $0.000$ & $\mathbf{0.000}$ & $0.000$ & $0.000$ \\
    	toxicity & $0.783$ & $\mathbf{0.049}$ & $0.365$ & $0.128$ & $0.281$ & $0.001$ & $0.275$ & $0.001$ & $0.000$ \\
    	yacht & $0.135$ & $\mathbf{0.000}$ & $\mathbf{0.000}$ & $\mathbf{0.000}$ & $0.000$ & $0.000$ & $\mathbf{0.001}$ & $0.000$ & $0.000$ \\
    \hline 
    \end{tabular}
    \caption{P-values returned by the Wilcoxon rank-sum test under the alternative hypothesis that the median errors on the test set obtained from classical GSGP are equal with respect to the errors obtained with the methods introduced in this paper. Highlighted in bold, the p-values below $0.05$ where the direction of the difference shows an improvement with respect to standard GSGP.}
    \label{tab:p-values}
\end{table}

\begin{figure}[tp]
    \centering
    \subfigure[air]{\includegraphics[width=.48\textwidth]{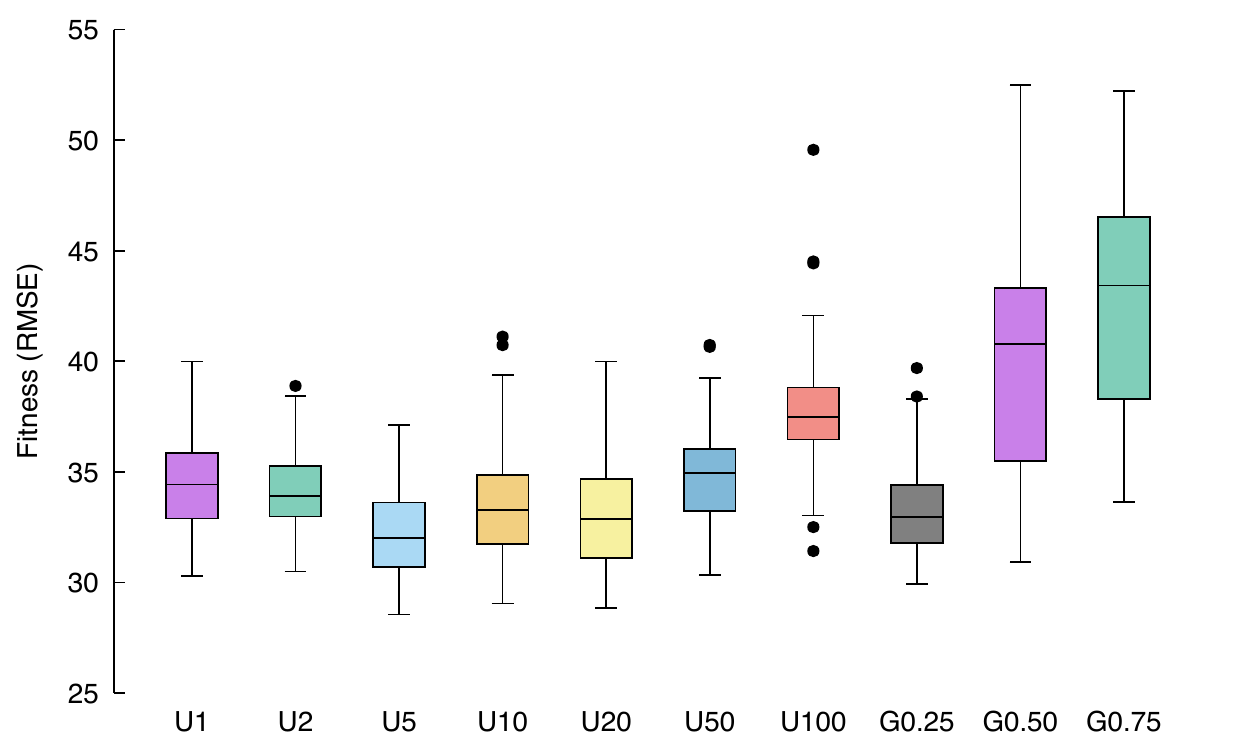}}  
    \subfigure[\%F]{\includegraphics[width=.48\textwidth]{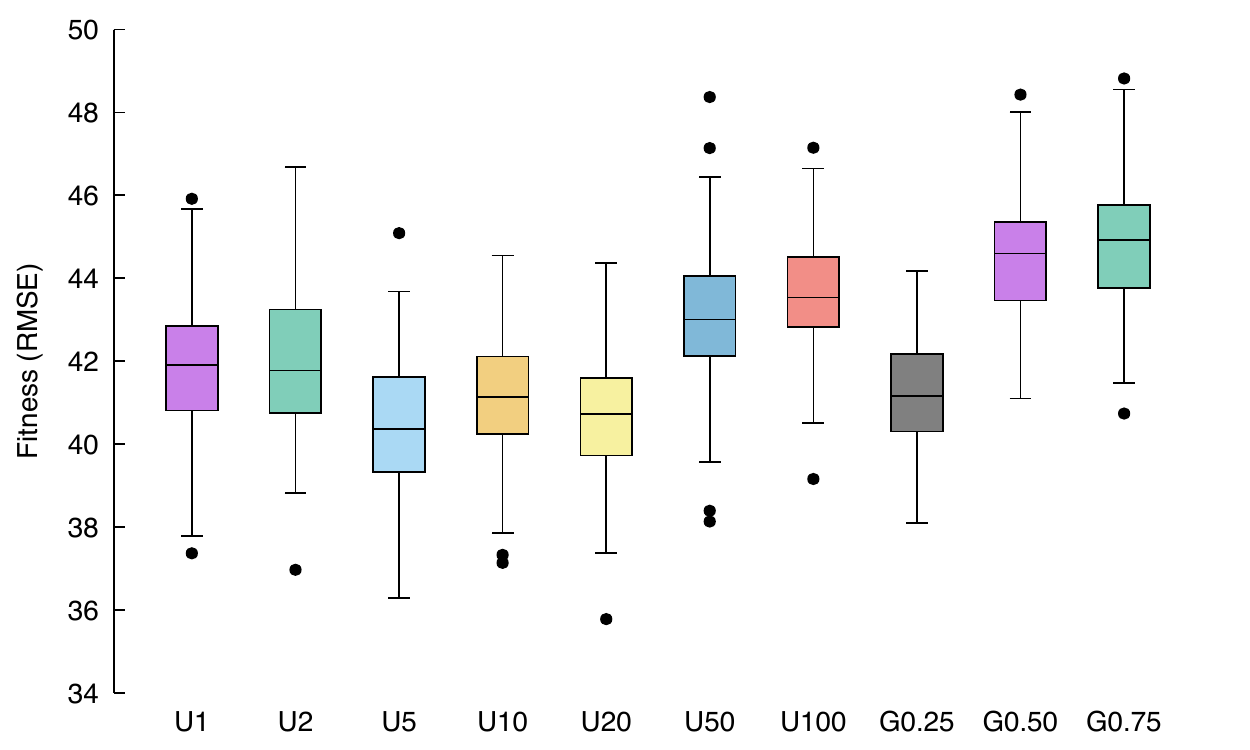}}   
    \subfigure[conc]{\includegraphics[width=.48\textwidth]{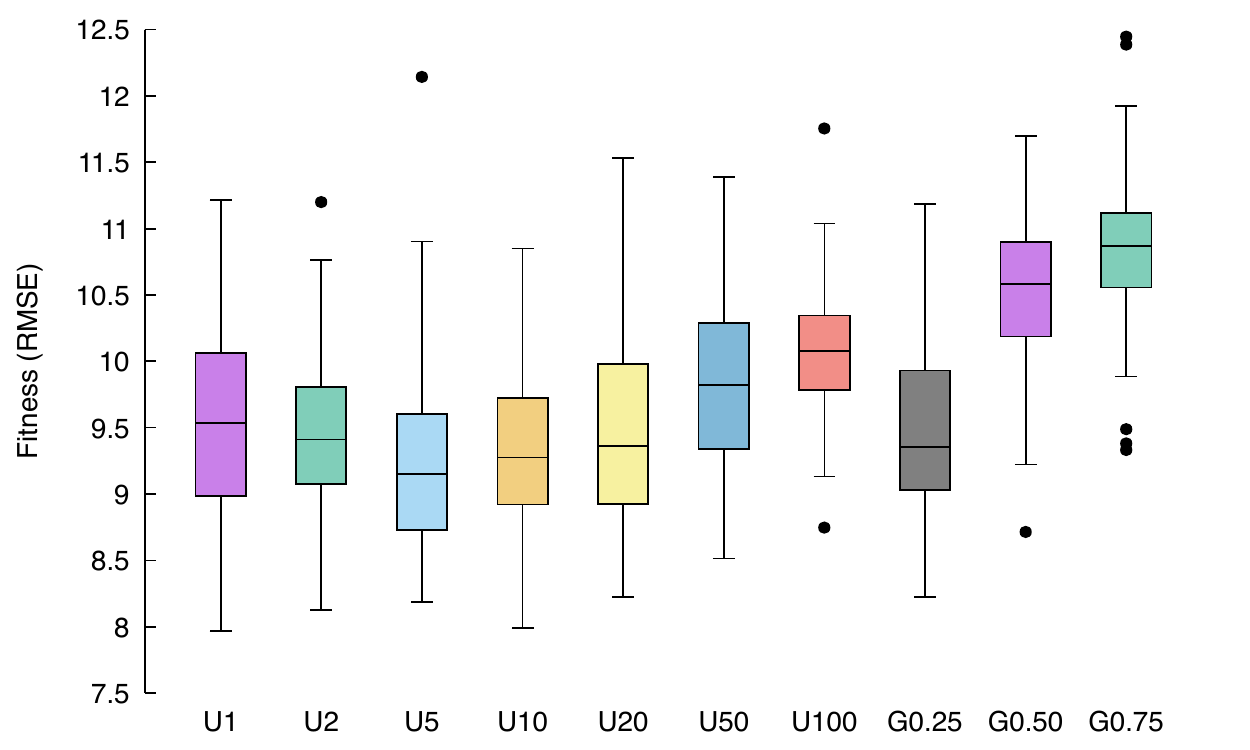}}
    \subfigure[\%PPB]{\includegraphics[width=.48\textwidth]{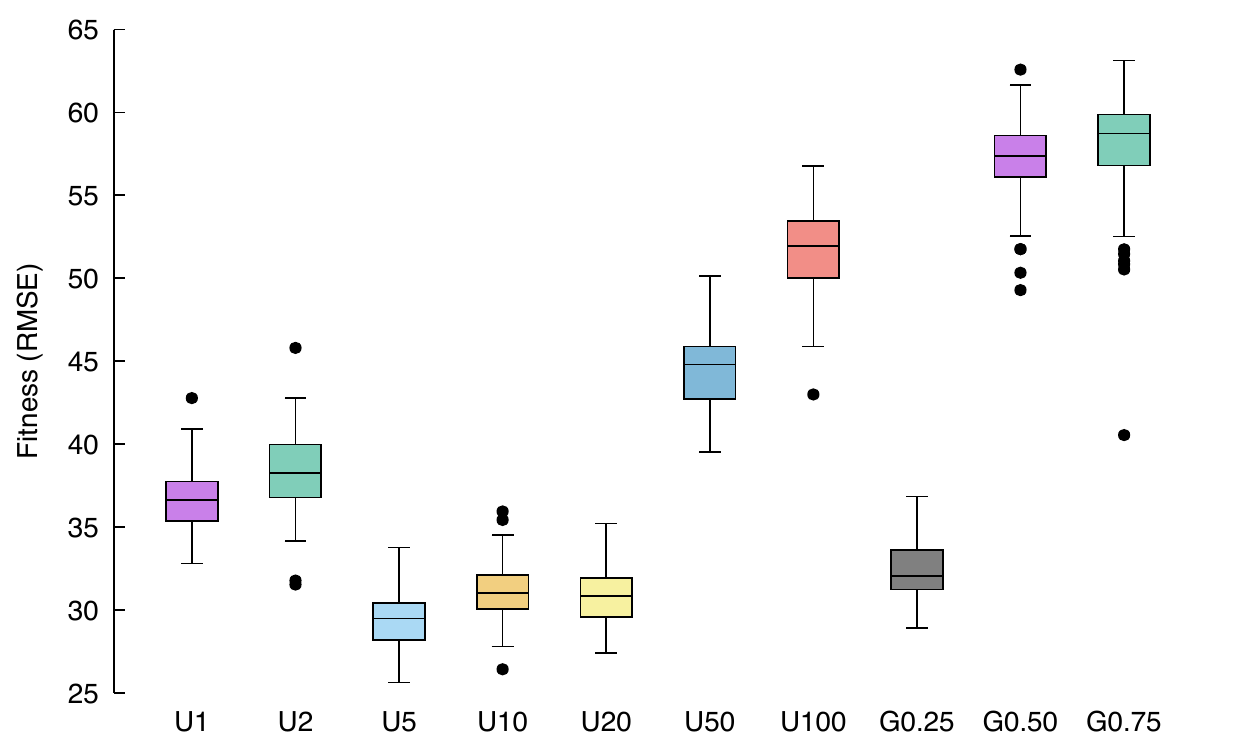}} 
    \subfigure[LD50]{\includegraphics[width=.48\textwidth]{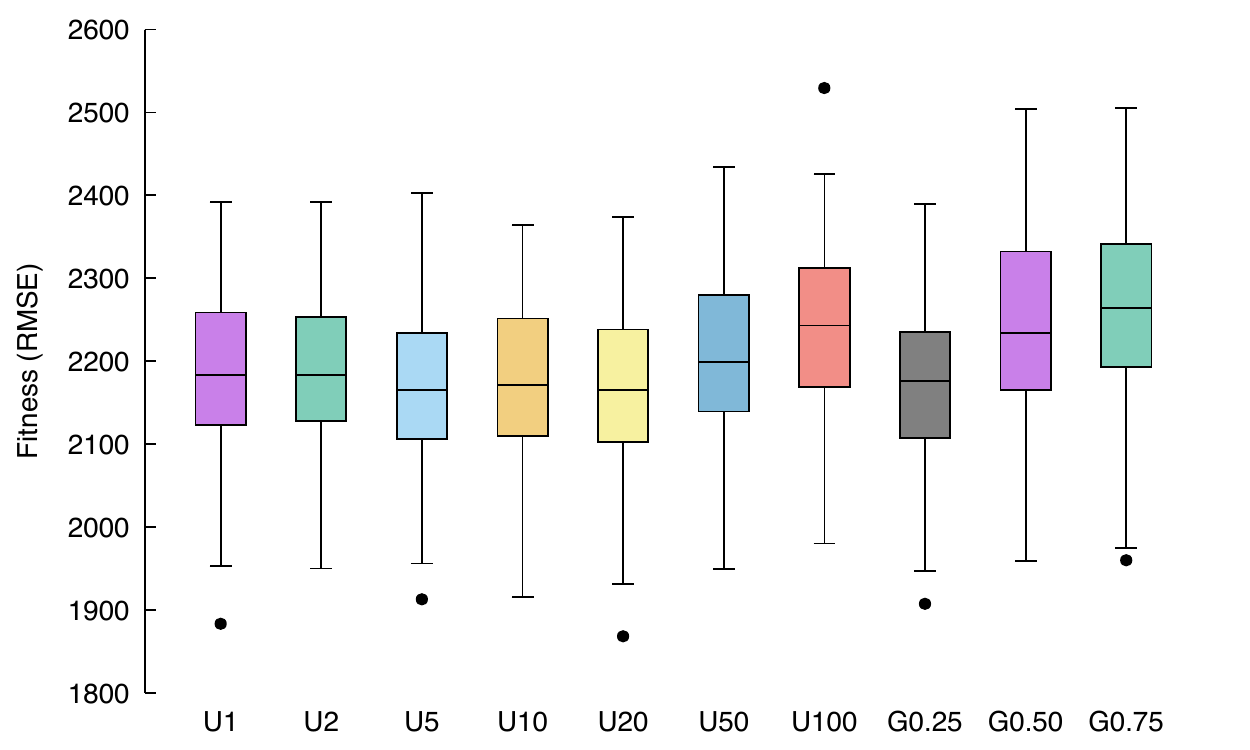}}
    \subfigure[yac]{\includegraphics[width=.48\textwidth]{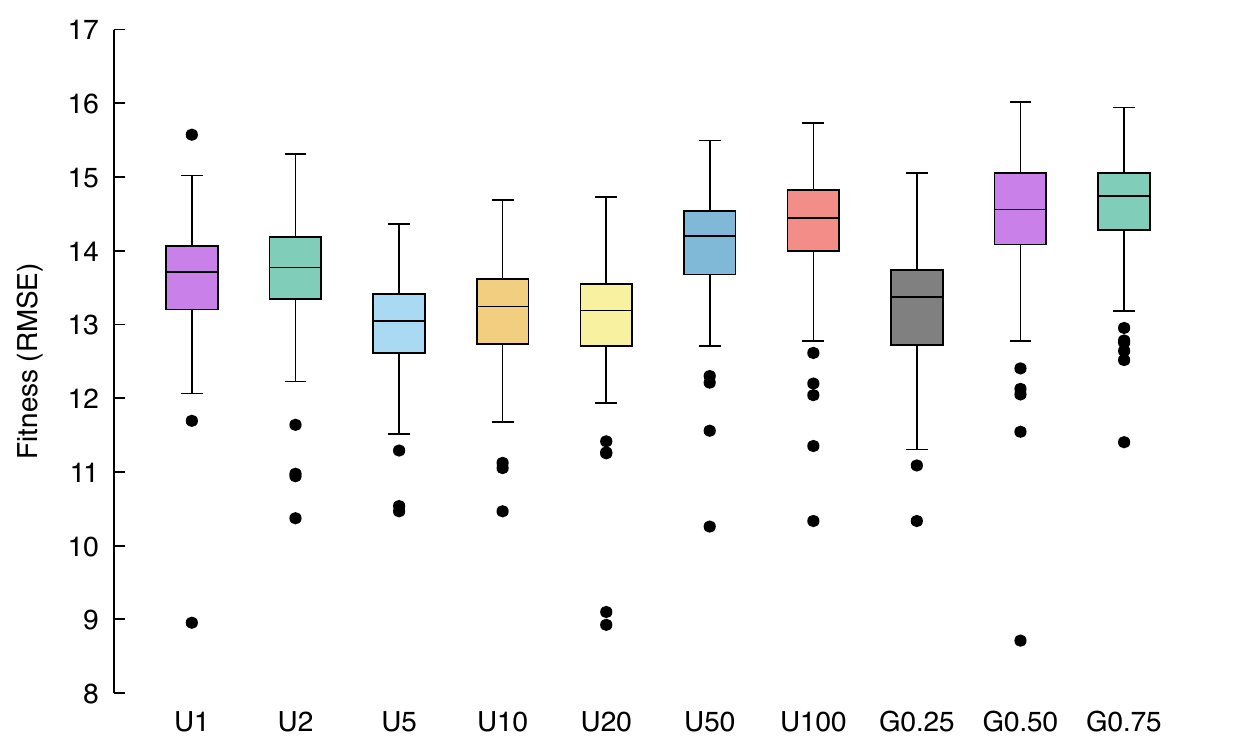}}
    \caption{Box-plots of the RMSE on the training set over $100$ independent runs of the considered benchmark dataset for all the proposed methods.}
    \label{fig:bp-train}
\end{figure}

\begin{figure}[tp]
    \centering
    \subfigure[air]{\includegraphics[width=.48\textwidth]{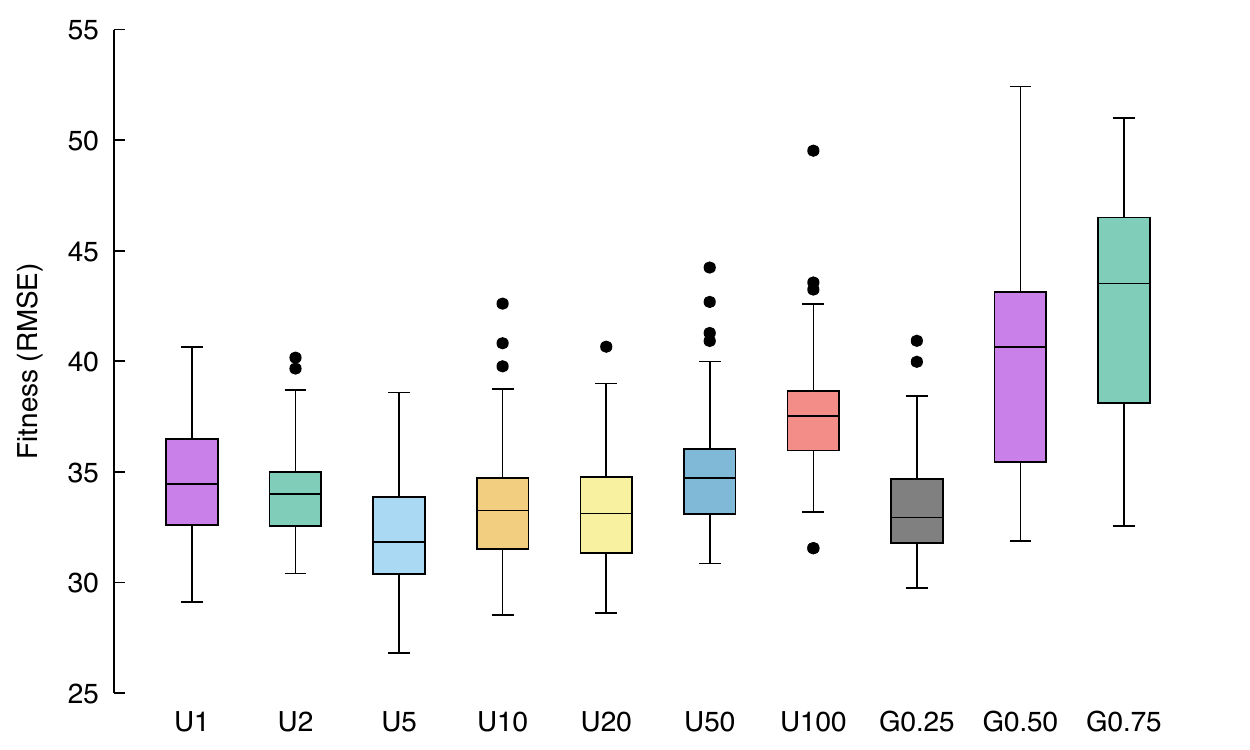}}   
    \subfigure[\%F]{\includegraphics[width=.48\textwidth]{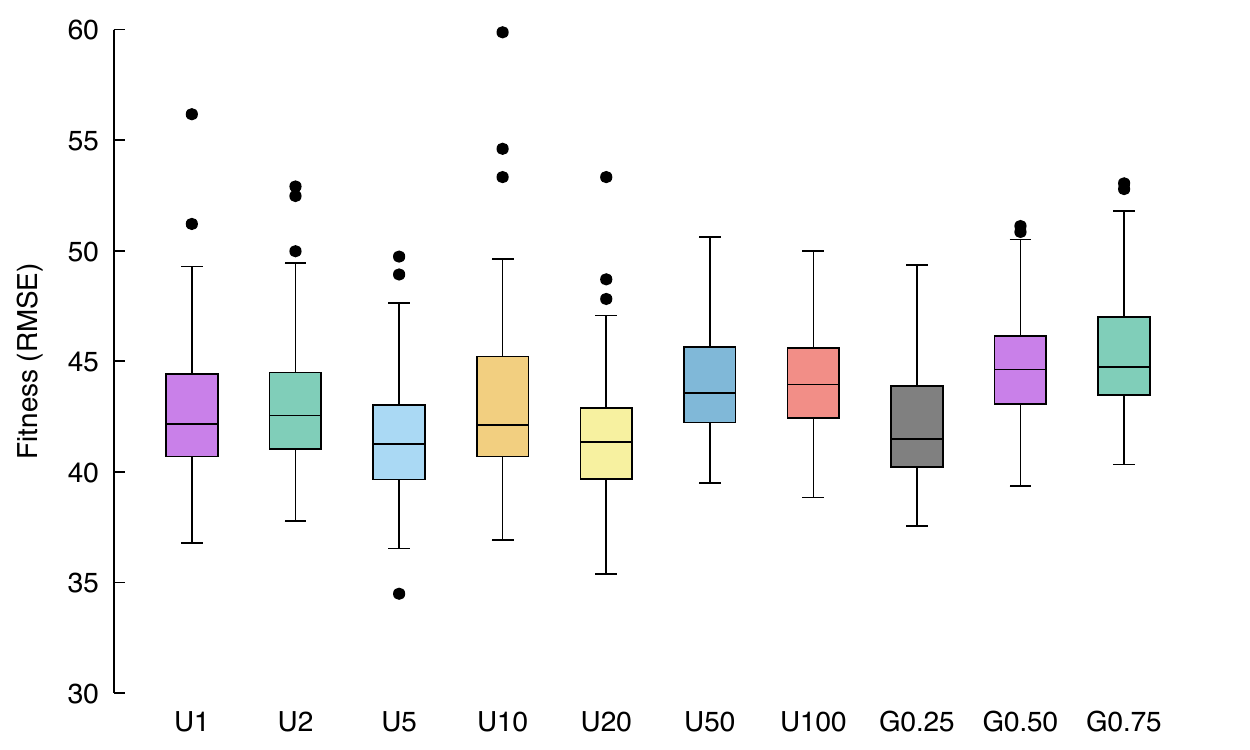}}   
    \subfigure[conc]{\includegraphics[width=.48\textwidth]{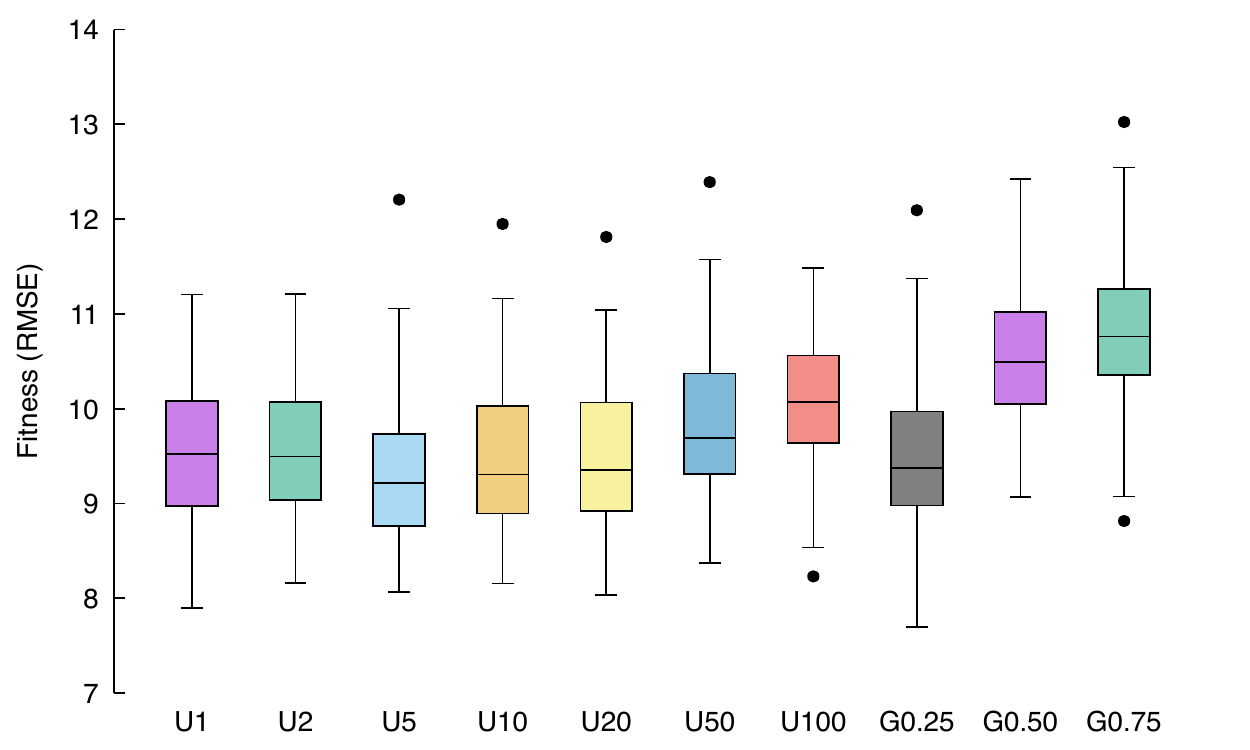}}
    \subfigure[\%PPB]{\includegraphics[width=.48\textwidth]{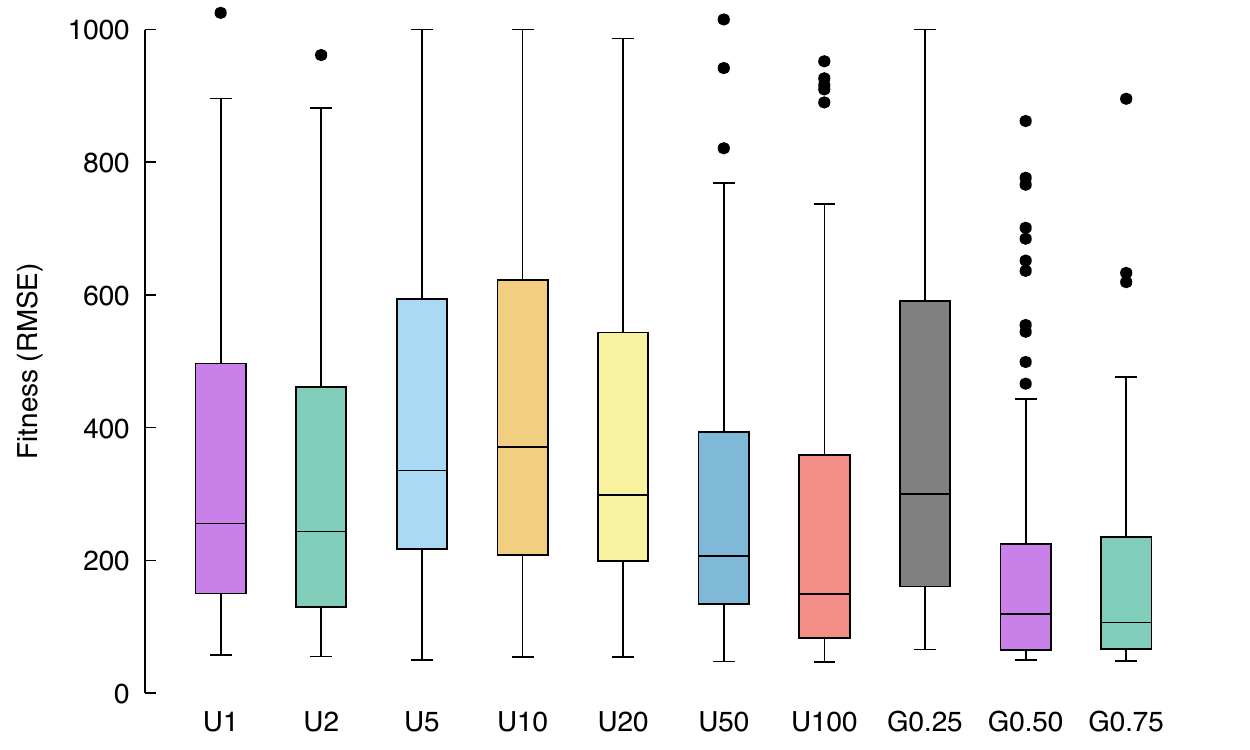}} 
    \subfigure[LD50]{\includegraphics[width=.48\textwidth]{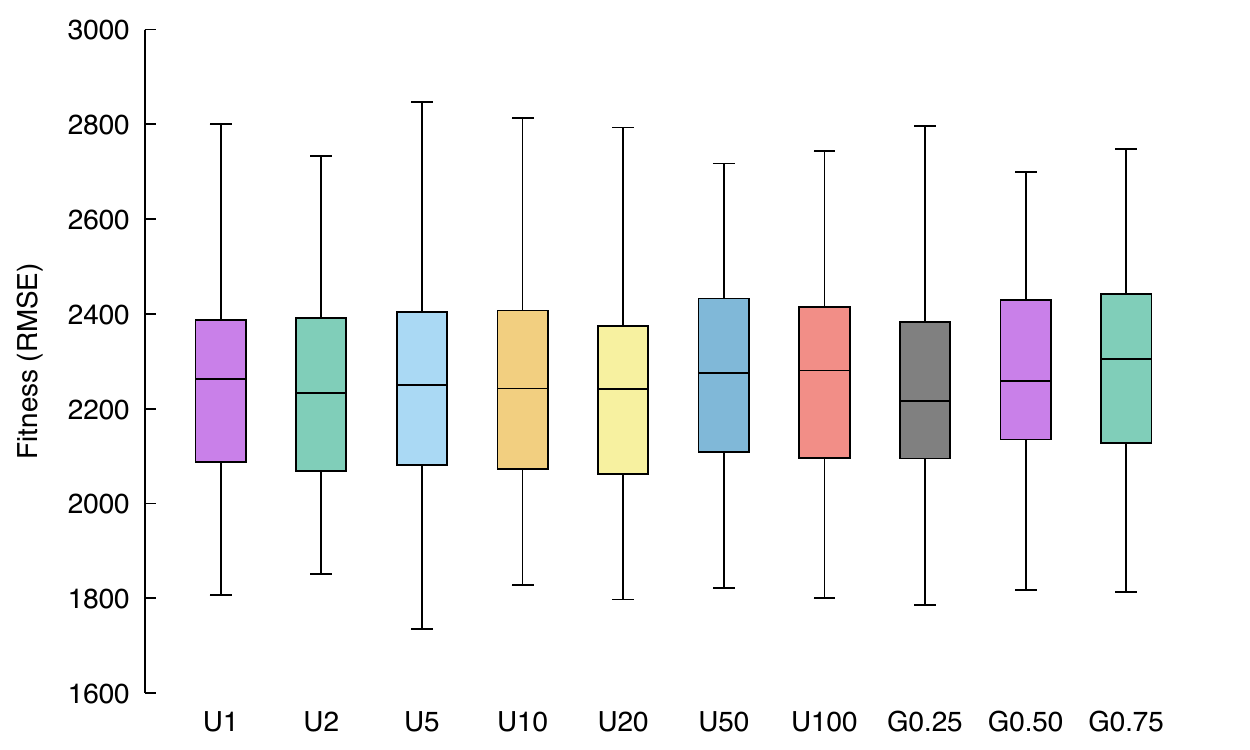}} 
    \subfigure[yac]{\includegraphics[width=.48\textwidth]{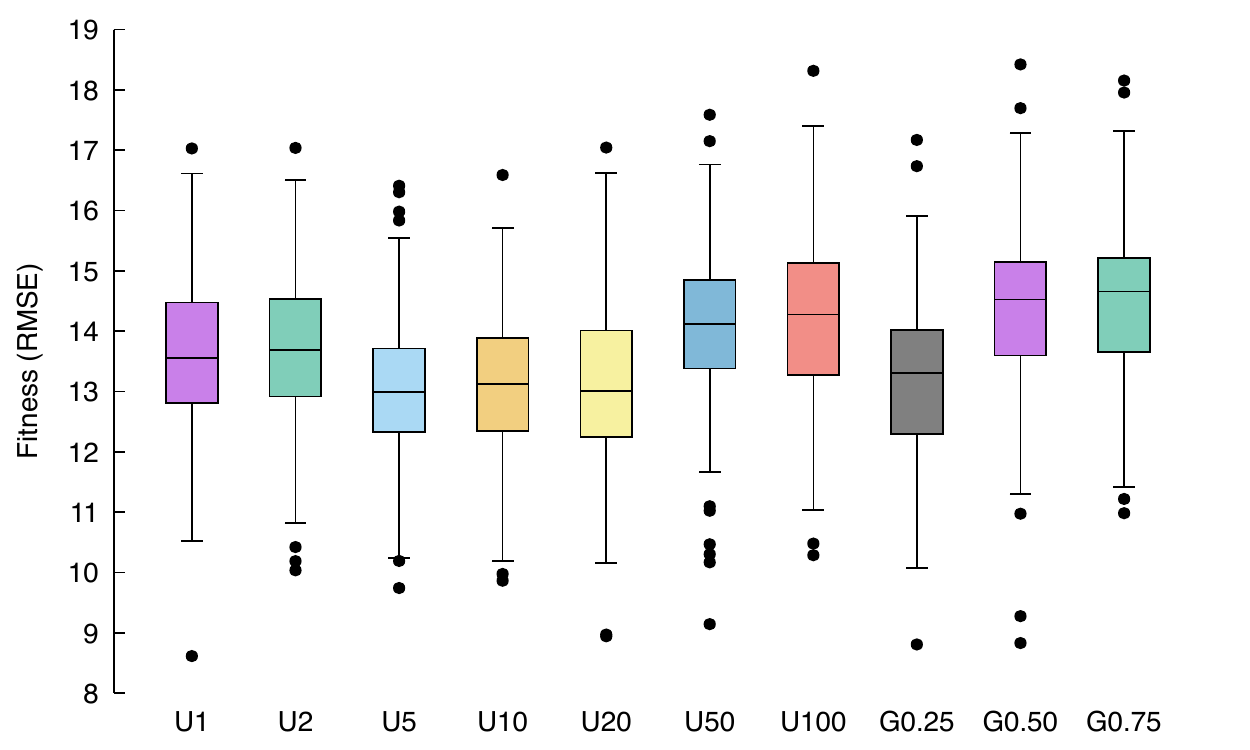}}
    \caption{Box-plots of the RMSE on the test set over $100$ independent runs of the considered benchmark dataset for all the proposed methods.}
    \label{fig:bp-test}
\end{figure}

Figures~\ref{fig:bp-train} and~\ref{fig:bp-test} show, via box-plots, the distribution of the fitness, calculated over $100$ independent run, achieved by the different configurations, discussed in Section~\ref{subsec:experimental-study}, compared to classical GSGP, which is denoted here as U$1$ whereas, as explained before, it considers only the previous generation from which to select parents, exactly as in GSGP.
Table~\ref{tab:uniform} stores the fitness values obtained by selecting the ancestors with Uniform Multi-Generational
Selection, for both training and testing set among all the considered benchmark problems. 
Similarly, Table~\ref{tab:geometric} stores Geometric Multi-Generational selection method fitness values. 
Table~\ref{tab:p-values} reports a statistical significance assessment of the achieved result, displaying the \emph{p}-values obtained under the hypothesis that the median fitness resulting from the considered technique is equal to the one obtained with standard GSGP. This means that, if the resulting \emph{p}-values is zero, the fitness obtained is statistically significantly better or significantly worse with respect to the one achieved by GSGP, and to understand which case we are dealing with, the results contained in Table~\ref{tab:uniform} and Table~\ref{tab:geometric} must be compared.
Finally, Table~\ref{tab:uniform} and Table~\ref{tab:geometric} display the fitness values obtained by selecting parents using a uniform distribution and geometric distribution respectively.  

Concerning the \emph{air} dataset (Figures~\ref{fig:bp-train}(a) and~\ref{fig:bp-test}(a)) it is possible to see that, if we consider a uniform distribution for the multi-generational selection, the best performance is obtained by selecting ancestors from the $5$ previous generations (Table~\ref{tab:uniform}); on the other hand, with a geometric distribution better results are achieved with $p=0.25$ (Table~\ref{tab:geometric}). Both of these methods meaningfully outperform standard GSGP, as confirmed by the statistical results of Table~\ref{tab:p-values}.
For what concerns the other variations of the proposed methods, combining information provided by fitness distribution shown in the box plots with the corresponding \emph{p}-values, it is possible to conclude that: U$2$, U$10$ and U$20$ also outperform GSGP with statistical significance, U$50$ leads to results similar to GSGP, while U$100$, G$0.50$ and G$0.75$ results are substantially less performing w.r.t. to GSGP.

Considering the \%F dataset (Figures~\ref{fig:bp-train}(b) and~\ref{fig:bp-test}(b)), a similar behavior as the one observed for the previous benchmark problem appears. Again, U$5$ and G$0.25$ represent the two best candidates and outperform GSGP together with U$2$, U$10$, U$20$. Here, U$50$, U$100$, G$0.50$, and G$0.75$ lead to worst fitness values compared to classical GSGP.

Moving to the \emph{conc} dataset, (Figures~\ref{fig:bp-train}(c) and~\ref{fig:bp-test}(c) provides us with more evidence that U$5$ and G$0.25$ are a significant improvement of the standard GSGP. Moreover, it is still the case that U$2$, U$10$ and U$20$ slightly outperform GSGP, while U$50$, U$100$, G$0.50$ and G$0.75$ bring poor fitness values.

Taking into account the \%PPB dataset (Figures~\ref{fig:bp-train}(d) and~\ref{fig:bp-test}(d)), it is clear that all models are affected by overfitting, and a lower error in the training set entails a bigger error in the test set. 

Applying our methods to the LD50 dataset (Figures~\ref{fig:bp-train}(e) and~\ref{fig:bp-test}(e)) we can once more recognize that U$5$ and G$0.25$ represent the best improvement of GSGP. Further, it is important to highlight that for this dataset, while U$2$, U$10$, U$20$, U$50$, U$100$, G$0.50$ and G$0.75$ are indeed less performing than the standard GSGP, none of these methods results in actually poor fitness values, instead each of them reaches performance very similar to the one obtained by GSGP. 

Finally, the \emph{yac} dataset (Figures~\ref{fig:bp-train}(f) and~\ref{fig:bp-test}(f)) is the last confirmation of the behavior we have been observing so far. Once more, U$5$ and G$0.25$ are a significant improvement of the standard GSGP, U$2$, U$10$ and U$20$ outperform GSGP, whereas U$50$, U$100$, G$0.50$ and G$0.75$ result in poor fitness values. 

All in all, experiments performed revealed that U$5$ and G$0.25$ provide us with a meaningful improvement of the standard GSGP.

\begin{table}
\scriptsize
\centering
\begin{tabular}{rccccccccc|}
\multicolumn{2}{l}{}                           &       & GSGP     & U2                & U5                & U10      & U20      & U50      & U100             \\ \hline
\multicolumn{2}{r|}{\multirow{2}{*}{air}}   & train & 34.43   & 33.89            & \textbf{32.02}   & 33.28   & 32.85   & 34.93   & 34.93           \\ \cline{3-10} 
\multicolumn{2}{r|}{}                          & test  & 34.44   & 34.01            & \textbf{31.83}   & 33.26   & 33.12   & 34.72   & 37.51           \\ \hline
\multicolumn{2}{r|}{\multirow{2}{*}{\%F}}    & train & 41.92   & 41.78            & \textbf{40.37}   & 41.13   & 40.72   & 43.00   & 43.53           \\ \cline{3-10} 
\multicolumn{2}{r|}{}                          & test  & 42.17   & 42.54            & \textbf{41.27}   & 42.10   & 41.36   & 43.56   & 43.94           \\ \hline
\multicolumn{2}{r|}{\multirow{2}{*}{conc}} & train & 9.54    & 9.41             & \textbf{9.15}    & 9.27    & 9.36    & 9.82    & 10.08           \\ \cline{3-10} 
\multicolumn{2}{r|}{}                          & test  & 9.52    & 9.49             & \textbf{9.21}    & 9.31    & 9.35    & 9.69    & 10.07           \\ \hline
\multicolumn{2}{r|}{\multirow{2}{*}{\%PPB}}      & train & 36.62   & 38.26            & \textbf{29.48}   & 31.02   & 30.82   & 44.80   & 51.94           \\ \cline{3-10} 
\multicolumn{2}{r|}{}                          & test  & 255.51  & 243.46           & 335.42           & 371.03  & 298.03  & 206.88  & \textbf{148.83} \\ \hline
\multicolumn{2}{r|}{\multirow{2}{*}{LD50}} & train & 2183.65 & 2183.17          & \textbf{2165.20} & 2171.09 & 2165.36 & 2199.38 & 2243.06         \\ \cline{3-10} 
\multicolumn{2}{r|}{}                          & test  & 2262.15 & \textbf{2233.41} & 2250.19          & 2242.84 & 2240.93 & 2274.87 & 2280.51         \\ \hline
\multicolumn{2}{r|}{\multirow{2}{*}{yac}}    & train & 13.71   & 13.77            & \textbf{13.04}   & 13.24   & 13.18   & 14.19   & 14.44           \\ \cline{3-10} 
\multicolumn{2}{r|}{}                          & test  & 13.55   & 13.69            & \textbf{12.99}   & 13.12   & 13.00   & 14.11   & 14.27           \\ \hline
\end{tabular}
\caption{Fitness values obtained  by selecting the ancestors with Uniform Multi-Generational Selection. The values in bold are the best results obtained.}
    \label{tab:uniform}
\end{table}

Regarding results obtained with Uniform Multi-Generational Selection, a number $k$ of previous generations equal to $10$ and $20$ (and $5$, as aforementioned) lead to statistically significant improvement in the fitness. This confirms our intuition: selecting ancestors also from previous generations (that, anyway, are not too far away) led to better results as a more wide set of genotypes is considered for recombination, and good characteristics of an individual that may have been lost during generation can be retrieved, thus decreasing the likelihood of being stuck in local minima.
On the other hand, considering only $2$ previous generations results in fitness values comparable with GSGP. This is reasonable considering that individuals of two subsequent generations do not differ too much. Thus, selecting ancestors from a generation or from the directly previous one does not remarkably affect the quality of the offspring generated. 
On the other hand, U$50$ and U$100$ lead to significantly worse performance in terms of fitness. This is because ancestors are selected from generations too far back, where individuals were not yet improved by the genetic process.

\begin{table}
\small
\centering
\begin{tabular}{rcccccc|}
\multicolumn{2}{l}{}                           &    & GSGP       & G0.25             & G0.50           & G0.75            \\ \hline
\multicolumn{2}{r|}{\multirow{2}{*}{air}}  & train & 34.43 & 37.48            & \textbf{32.97} & 40.76           \\ \cline{3-7} 
\multicolumn{2}{r|}{}                          & test & 34.44  & \textbf{32.93}   & 40.63          & 43.51           \\ \hline
\multicolumn{2}{r|}{\multirow{2}{*}{\%F}}    & train & 41.92 & \textbf{41.16}   & 44.60          & 44.92           \\ \cline{3-7} 
\multicolumn{2}{r|}{}                          & test & 42.17  & \textbf{41.49}   & 44.63          & 44.73           \\ \hline
\multicolumn{2}{r|}{\multirow{2}{*}{conc}} & train & 9.54 & \textbf{9.35}    & 10.58          & 10.86           \\ \cline{3-7} 
\multicolumn{2}{r|}{}                          & test & 9.52  & \textbf{9.37}    & 10.48          & 10.76           \\ \hline
\multicolumn{2}{r|}{\multirow{2}{*}{\%PPB}}      & train & 36.62 & \textbf{32.06}   & 57.37          & 58.73           \\ \cline{3-7} 
\multicolumn{2}{r|}{}                          & test & 255.51  & 300.03           & 119.20         & \textbf{106.33} \\ \hline
\multicolumn{2}{r|}{\multirow{2}{*}{LD50}} & train & 2183.65 & \textbf{2176.51} & 2234.56        & 2264.43         \\ \cline{3-7} 
\multicolumn{2}{r|}{}                          & test & 2262.15  & \textbf{2216.47} & 2258.10        & 2305.45         \\ \hline
\multicolumn{2}{r|}{\multirow{2}{*}{yac}}    & train & 13.71 & \textbf{13.37}   & 14.55          & 14.73           \\ \cline{3-7} 
\multicolumn{2}{r|}{}                          & test & 13.55  & \textbf{13.30}   & 14.52          & 14.65           \\ \hline
\end{tabular}
\caption{Fitness values obtained  by selecting the ancestors with Geometric Multi-Generational Selection. The values in bold are the best results obtained.}
    \label{tab:geometric}
\end{table}

Considering results achieved by Geometric Multi-Generational Selection, while, as stated above, setting $p=0.25$ led to a significant upgrade of the fitness value, for the other choice of parameters, i.e. $p=0.5$ and $p=0.75$, the obtained results worse w.r.t. standard GSGP. 

These results are interesting for a particular reason: the expected value of the geometric distributions with $p=0.25$, $p=0.5$, and $p=0.75$ are $3$, $1$, and $\frac{1}{3}$, respectively (i.e., $\frac{1-p}{p}$). Thus, we would expect G$0.25$, G$0.50$, and G$0.75$ to behave similarly to U$4$, U$2$, and between U$1$ and U$2$, respectively. In the first case, it appears to be correct, while in the second one this happens only in some of the datasets. However, G$0.75$ behaviour is quite different from what was expected. Since the motivation cannot be traced back to the expected value of the distribution, it could be due the fact that while in expectation most of the individuals will be from the previous generation, only a limited number of them can be from older ones, damaging the search process. However, this is only a conjecture, and we expect to investigate this unexpected behaviour in later works, together with the effect of using other distributions in the multi-generational selection.

\section{Conclusions}
\label{sec:conclusions}

In this paper, we have presented and studied a way to use the information of the previous generations, which GSGP stores anyway, to improve the performance of GSGP. This resulted in a multi-generational selection, and we presented two methods implementing this idea: a uniform selection probability and a geometric selection. The main idea consists in selecting individuals not only from the last population but also from previous ones (how many and with which probability depending on the underlying method). We have tested the proposed uniform and geometric multi-generational selections with multiple parameters on a selection of six datasets, showing that a \emph{limited} ability to select from previous populations is beneficial to the search process. In the future, we plan to expand this research and provide even more powerful ways of exploiting the additional information that GSGP, in its fast implementation, is already storing.

\bibliographystyle{abbrv}
\bibliography{bibliography}

\begin{thebibliography}{10}

\bibitem{archetti2006genetic}
F.~Archetti, S.~Lanzeni, E.~Messina, and L.~Vanneschi.
\newblock Genetic programming for human oral bioavailability of drugs.
\newblock In {\em Proceedings of the 8th annual conference on Genetic and
  Evolutionary Computation}, pages 255--262, 2006.

\bibitem{archetti2007genetic}
F.~Archetti, S.~Lanzeni, E.~Messina, and L.~Vanneschi.
\newblock Genetic programming and other machine learning approaches to predict
  median oral lethal dose (ld 50) and plasma protein binding levels (\% ppb) of
  drugs.
\newblock In {\em European Conference on Evolutionary Computation, Machine
  Learning and Data Mining in Bioinformatics}, pages 11--23. Springer, 2007.

\bibitem{augusto00}
D.~A. Augusto and H.~J.~C. Barbosa.
\newblock Symbolic regression via genetic programming.
\newblock In F.~M.~G. Fran{\c{c}}a and C.~H.~C. Ribeiro, editors, {\em 6th
  Brazilian Symposium on Neural Networks {(SBRN} 2000), 22-25 November 2000,
  Rio de Janiero, Brazil}, pages 173--178. {IEEE} Computer Society, 2000.

\bibitem{bi21}
Y.~Bi, B.~Xue, and M.~Zhang.
\newblock A divide-and-conquer genetic programming algorithm with ensembles for
  image classification.
\newblock {\em {IEEE} Trans. Evol. Comput.}, 25(6):1148--1162, 2021.

\bibitem{brooks1989airfoil}
T.~F. Brooks, D.~S. Pope, and M.~A. Marcolini.
\newblock Airfoil self-noise and prediction.
\newblock Technical report, 1989.

\bibitem{cao09}
Y.~Cao and W.~Luo.
\newblock Novel associative memory retrieving strategies for evolutionary
  algorithms in dynamic environments.
\newblock In Z.~Cai, Z.~Li, Z.~Kang, and Y.~Liu, editors, {\em Advances in
  Computation and Intelligence, 4th International Symposium, {ISICA} 2009,
  Huangshi, China, Ocotober 23-25, 2009, Proceedings}, volume 5821 of {\em
  Lecture Notes in Computer Science}, pages 258--268. Springer, 2009.

\bibitem{castelli2019gsgp}
M.~Castelli and L.~Manzoni.
\newblock Gsgp-c++ 2.0: A geometric semantic genetic programming framework.
\newblock {\em SoftwareX}, 10:100313, 2019.

\bibitem{castelli19}
M.~Castelli, L.~Manzoni, L.~Mariot, and M.~Saletta.
\newblock Extending local search in geometric semantic genetic programming.
\newblock In P.~M. Oliveira, P.~Novais, and L.~P. Reis, editors, {\em Progress
  in Artificial Intelligence - 19th {EPIA} Conference on Artificial
  Intelligence, {EPIA} 2019, Vila Real, Portugal, September 3-6, 2019,
  Proceedings, Part {I}}, volume 11804 of {\em Lecture Notes in Computer
  Science}, pages 775--787. Springer, 2019.

\bibitem{castelli2017influence}
M.~Castelli, L.~Manzoni, S.~Silva, L.~Vanneschi, and A.~Popovi{\v{c}}.
\newblock The influence of population size in geometric semantic gp.
\newblock {\em Swarm and Evolutionary Computation}, 32:110--120, 2017.

\bibitem{castelli2011effect}
M.~Castelli, L.~Manzoni, and L.~Vanneschi.
\newblock The effect of selection from old populations in genetic algorithms.
\newblock In N.~Krasnogor and P.~L. Lanzi, editors, {\em 13th Annual Genetic
  and Evolutionary Computation Conference, {GECCO} 2011, Companion Material
  Proceedings, Dublin, Ireland, July 12-16, 2011}, pages 161--162. {ACM}, 2011.

\bibitem{castelli2011method}
M.~Castelli, L.~Manzoni, and L.~Vanneschi.
\newblock A method to reuse old populations in genetic algorithms.
\newblock In L.~Antunes and H.~S. Pinto, editors, {\em Progress in Artificial
  Intelligence, 15th Portuguese Conference on Artificial Intelligence, {EPIA}
  2011, Lisbon, Portugal, October 10-13, 2011. Proceedings}, volume 7026 of
  {\em Lecture Notes in Computer Science}, pages 138--152. Springer, 2011.

\bibitem{castelli15}
M.~Castelli, L.~Trujillo, L.~Vanneschi, S.~Silva, E.~Z.{-}Flores, and
  P.~Legrand.
\newblock Geometric semantic genetic programming with local search.
\newblock In S.~Silva and A.~I. Esparcia{-}Alc{\'{a}}zar, editors, {\em
  Proceedings of the Genetic and Evolutionary Computation Conference, {GECCO}
  2015, Madrid, Spain, July 11-15, 2015}, pages 999--1006. {ACM}, 2015.

\bibitem{castelli2013prediction}
M.~Castelli, L.~Vanneschi, and S.~Silva.
\newblock Prediction of high performance concrete strength using genetic
  programming with geometric semantic genetic operators.
\newblock {\em Expert Systems with Applications}, 40(17):6856--6862, 2013.

\bibitem{jaskowski07}
W.~Jaskowski, K.~Krawiec, and B.~Wieloch.
\newblock Knowledge reuse in genetic programming applied to visual learning.
\newblock In H.~Lipson, editor, {\em Genetic and Evolutionary Computation
  Conference, {GECCO} 2007, Proceedings, London, England, UK, July 7-11, 2007},
  pages 1790--1797. {ACM}, 2007.

\bibitem{koza1994genetic}
J.~R. Koza.
\newblock Genetic programming as a means for programming computers by natural
  selection.
\newblock {\em Statistics and computing}, 4(2):87--112, 1994.

\bibitem{louis97}
S.~Louis and G.~Li.
\newblock Augmenting genetic algorithms with memory to solve traveling salesman
  problems.
\newblock In {\em Proceedings of the Joint Conference on Information Sciences},
  pages 108--111, 1997.

\bibitem{mcdermott2012genetic}
J.~McDermott, D.~R. White, S.~Luke, L.~Manzoni, M.~Castelli, L.~Vanneschi,
  W.~Jaskowski, K.~Krawiec, R.~Harper, K.~De~Jong, et~al.
\newblock Genetic programming needs better benchmarks.
\newblock In {\em Proceedings of the 14th annual conference on Genetic and
  evolutionary computation}, pages 791--798, 2012.

\bibitem{moraglio2012geometric}
A.~Moraglio, K.~Krawiec, and C.~G. Johnson.
\newblock Geometric semantic genetic programming.
\newblock In {\em International Conference on Parallel Problem Solving from
  Nature}, pages 21--31. Springer, 2012.

\bibitem{moraglio2004topological}
A.~Moraglio and R.~Poli.
\newblock Topological interpretation of crossover.
\newblock In {\em Genetic and Evolutionary Computation Conference}, pages
  1377--1388. Springer, 2004.

\bibitem{pei19}
W.~Pei, B.~Xue, L.~Shang, and M.~Zhang.
\newblock Reuse of program trees in genetic programming with a new fitness
  function in high-dimensional unbalanced classification.
\newblock In M.~L{\'{o}}pez{-}Ib{\'{a}}{\~{n}}ez, A.~Auger, and
  T.~St{\"{u}}tzle, editors, {\em Proceedings of the Genetic and Evolutionary
  Computation Conference Companion, {GECCO} 2019, Prague, Czech Republic, July
  13-17, 2019}, pages 187--188. {ACM}, 2019.

\bibitem{pietropolli2022combining}
G.~Pietropolli, L.~Manzoni, A.~Paoletti, and M.~Castelli.
\newblock Combining geometric semantic gp with gradient-descent optimization.
\newblock In {\em European Conference on Genetic Programming (Part of
  EvoStar)}, pages 19--33. Springer, 2022.

\bibitem{seront95}
G.~Seront.
\newblock External concepts reuse in genetic programming.
\newblock In {\em working notes for the AAAI Symposium on Genetic programming},
  pages 94--98. MIT/AAAI Cambridge, 1995.

\bibitem{sipper20}
M.~Sipper and J.~H. Moore.
\newblock Conservation machine learning.
\newblock {\em BioData Min.}, 13(1):9, 2020.

\bibitem{sipper21}
M.~Sipper and J.~H. Moore.
\newblock Conservation machine learning: a case study of random forests.
\newblock {\em Scientific Reports}, 11(1):1--6, 2021.

\bibitem{vanneschi2013new}
L.~Vanneschi, M.~Castelli, L.~Manzoni, and S.~Silva.
\newblock A new implementation of geometric semantic gp and its application to
  problems in pharmacokinetics.
\newblock In {\em European Conference on Genetic Programming}, pages 205--216.
  Springer, 2013.

\bibitem{vanneschi2014survey}
L.~Vanneschi, M.~Castelli, and S.~Silva.
\newblock A survey of semantic methods in genetic programming.
\newblock {\em Genetic Programming and Evolvable Machines}, 15(2):195--214,
  2014.

\bibitem{vanneschi2014geometric}
L.~Vanneschi, S.~Silva, M.~Castelli, and L.~Manzoni.
\newblock Geometric semantic genetic programming for real life applications.
\newblock In {\em Genetic programming theory and practice xi}, pages 191--209.
  Springer, 2014.

\bibitem{wiering04}
M.~Wiering.
\newblock Memory-based memetic algorithms.
\newblock In {\em Benelearn'04: Proceedings of the Thirteenth Belgian-Dutch
  Conference on Machine Learning}, pages 191--198, 2004.

\bibitem{yang08}
S.~Yang.
\newblock Genetic algorithms with memory- and elitism-based immigrants in
  dynamic environments.
\newblock {\em Evol. Comput.}, 16(3):385--416, 2008.

\end{thebibliography}

\end{document}